\def\isarxiv{1} 

\ifdefined\isarxiv
\documentclass[11pt]{article}

\usepackage[numbers]{natbib}

\else
\documentclass{article}
\usepackage{neurips_2022}
\fi

\usepackage{amsmath}
\usepackage{amsthm}
\usepackage{amssymb}
\usepackage{algorithm}
\usepackage{subfig}
\usepackage{algpseudocode}
\usepackage{graphicx}
\usepackage{grffile}
\usepackage{wrapfig,epsfig}
\usepackage{url}
\usepackage{xcolor}
\usepackage{epstopdf}

\usepackage{bbm}
\usepackage{dsfont}

\allowdisplaybreaks

\ifdefined\isarxiv

\usepackage{tikz}
\usepackage{hyperref}  
\hypersetup{colorlinks=true,citecolor=blue,linkcolor=blue} 
\usetikzlibrary{arrows}
\usepackage[margin=1in]{geometry}

\else

\usepackage{microtype}
\usepackage{hyperref}
\definecolor{mydarkblue}{rgb}{0,0.08,0.45}
\hypersetup{colorlinks=true, citecolor=mydarkblue,linkcolor=mydarkblue}

\fi

\newtheorem{theorem}{Theorem}[section]
\newtheorem{lemma}[theorem]{Lemma}
\newtheorem{definition}[theorem]{Definition}

\newtheorem{fact}[theorem]{Fact}

\newtheorem{claim}[theorem]{Claim}

\newcommand{\wh}{\widehat}
\newcommand{\wt}{\widetilde}

\newcommand{\R}{\mathbb{R}}

\renewcommand{\d}{\mathrm{d}}

\renewcommand{\hat}{\wh}

\DeclareMathOperator*{\E}{{\mathbb{E}}}

\DeclareMathOperator{\diag}{diag}

\makeatletter
\newcommand*{\RN}[1]{\expandafter\@slowromancap\romannumeral #1@}
\makeatother

\newcommand{\Tianyi}[1]{{\color{blue}[Tianyi: #1]}} 

\usepackage{lineno}

\begin{document}

\ifdefined\isarxiv

\date{}

\title{The Closeness of In-Context Learning and Weight Shifting for Softmax Regression}
\author{
Shuai Li\thanks{\texttt{shuaili8@sjtu.edu.cn}. Shanghai Jiao Tong University.}
\and
Zhao Song\thanks{\texttt{zsong@adobe.com}. Adobe Research.}
\and 
Yu Xia\thanks{\texttt{xiayuu@umich.edu}. University of Michigan.}
\and 
Tong Yu\thanks{\texttt{tyu@adobe.com}. Adobe Research.}
\and
Tianyi Zhou\thanks{\texttt{t8zhou@ucsd.edu}. University of California San Diego.}
}

\else

\title{Intern Project} 
\maketitle 
\fi

\ifdefined\isarxiv
\begin{titlepage}
  \maketitle
  \begin{abstract}

Large language models (LLMs) are known for their exceptional performance in natural language processing, making them highly effective in many human life-related or even job-related tasks. The attention mechanism in the Transformer architecture is a critical component of LLMs, as it allows the model to selectively focus on specific input parts. The softmax unit, which is a key part of the attention mechanism, normalizes the attention scores. Hence, the performance of LLMs in various NLP tasks depends significantly on the crucial role played by the attention mechanism with the softmax unit.

In-context learning, as one of the celebrated abilities of recent LLMs, is an important concept in querying LLMs such as ChatGPT. 
Without further parameter updates, Transformers can learn to predict based on few in-context examples. 
However, the reason why Transformers becomes in-context learners is not well understood.
Recently, several works \cite{asa+22,gtlv22,onr+22} have studied the in-context learning from a mathematical perspective based on a linear regression formulation $\min_x\| Ax  - b \|_2$,
which show Transformers' capability of learning linear functions in context.  

In this work, we study the in-context learning based on a softmax regression formulation $\min_{x} \| \langle \exp(Ax), {\bf 1}_n \rangle^{-1} \exp(Ax) - b \|_2$ of Transformer's attention mechanism. We show the upper bounds of the data transformations induced by a single self-attention layer and by gradient-descent on a $\ell_2$ regression loss for softmax prediction function,
which imply that when training self-attention-only Transformers for fundamental regression tasks, the models learned by gradient-descent and Transformers show great similarity. 

  \end{abstract}
  \thispagestyle{empty}
\end{titlepage}

{\hypersetup{linkcolor=black}
}
\newpage

\else

\begin{abstract}

Large language models (LLMs) are known for their exceptional performance in natural language processing, making them highly effective in many human life-related or even job-related tasks. The attention mechanism in the Transformer architecture is a critical component of LLMs, as it allows the model to selectively focus on specific input parts. The softmax unit, which is a key part of the attention mechanism, normalizes the attention scores. Hence, the performance of LLMs in various NLP tasks depends significantly on the crucial role played by the attention mechanism with the softmax unit.

In-context learning, as one of the celebrated abilities of recent LLMs, is an important concept in querying LLMs such as ChatGPT. 
Without further parameter updates, Transformers can learn to predict based on few in-context examples. 
However, the reason why Transformers becomes in-context learners is not well understood.
Recently, several works \cite{asa+22,gtlv22,onr+22} have studied the in-context learning from a mathematical perspective based on a linear regression formulation $\min_x\| Ax  - b \|_2$,
which show Transformers' capability of learning linear functions in context.  

In this work, we study the in-context learning based on a softmax regression formulation $\min_{x} \| \langle \exp(Ax), {\bf 1}_n \rangle^{-1} \exp(Ax) - b \|_2$ of Transformer's attention mechanism. We show the upper bounds of the data transformations induced by a single self-attention layer and by gradient-descent on a $\ell_2$ regression loss for softmax prediction function,
which imply that when training self-attention-only Transformers for fundamental regression tasks, the models learned by gradient-descent and Transformers show great similarity. 
\end{abstract}

\fi

\section{Introduction}

In recent years, there has been a significant increase in research and development in the field of Artificial Intelligence (AI), with large language models (LLMs) emerging as an effective way to tackle complex tasks. 
Transformers have achieved state-of-the-art results in various natural language processing tasks, such as machine translation \cite{pcr19,ghg+20}, language modeling, question answering, and text generation \cite{lsx+22}. As a result, they have become the preferred architecture for NLP.
Based on that architecture, BERT \cite{dclt18}, GPT-3 \cite{bmr+20}, PaLM \cite{cnd+22}, and OPT \cite{zrg+22} were proposed. They have demonstrated remarkable learning and reasoning capabilities and have proven to be more efficient than smaller models and traditional techniques when processing natural language. 
 
Additionally, LLMs can be fine-tuned for multiple purposes without requiring a new build from scratch, making them a versatile tool for AI applications.
A prime example of this is ChatGPT, a chat software developed by OpenAI utilizing GPT-3's potential to its fullest.
Moreover, GPT-4 \cite{openai23} has the capability to handle intricate tasks that its predecessors were unable to accomplish. It has demonstrated a remarkable level of proficiency comparable to human performance in various professional and academic benchmarks.

Transformers have a specific type of sequence-to-sequence neural network architecture. They utilize the attention mechanism \cite{vsp+17,rns+18,dclt18,bmr+20} that allows them to capture long-range dependencies and context from input data effectively. 
The core of the attention mechanism is the attention matrix which is comprised of rows and columns, corresponding to individual words or ``tokens''. The attention matrix represents the relationships within the given text. It 
measures the importance of each token in a sequence as it relates to the desired output.
During the training process, the attention matrix is learned and optimized to improve the accuracy of the model's predictions. Through the attention mechanism, each input token is evaluated based on its relevance to the desired output by assigning a token score. This score is determined by a similarity function that compares the current output state with input states.

Theoretically, the attention matrix is comprised of the query matrix  $Q \in \R^{n \times d}$, the key matrix $K \in \R^{n \times d}$ and the value matrix $V \in \R^{n \times d}$. Following \cite{zhdk23,as23,bsz23}, the computation of the normalized attention function is defined as $D^{-1} \exp(Q K^\top ) V$.  Following the transformer literature, we apply $\exp$ to a matrix entry-wise way. Here $D\in \R^{n \times n}$ is diagonal matrix that defined as $D = \diag( \exp(QK^\top ) {\bf 1}_n  ) $. Intuitively, $D$ denotes the softmax normalization  matrix. A more general computation formulation can be written as
\begin{align*}
\underbrace{ D^{-1} }_{n \times n \mathrm{~diagonal~matrix}} \underbrace{ \exp(X Q K^\top X^\top)}_{n \times n} \underbrace{ X }_{n \times d} \underbrace{ V }_{d \times d}, ~~~ D := \diag( \exp(X Q K^\top X^\top) {\bf 1}_n ) 
\end{align*}
In the above setting, we treat $Q, K, V\in \R^{d \times d}$ as weights and $X$ is the input sentence data that has length $n$ and each word embedding size is $d$. In the remaining of the part, we will switch $X$ to notation $A$ and use $A$ to denote sentence.

Mathematically, the attention computation problem can be formulated as a regression problem in the following sense
\begin{definition}\label{def:attention_regression}
We consider the following problem
\begin{align*}
\min_{X \in \R^{d \times d}} \| D^{-1} \exp( A X A^\top ) - B \|_F
\end{align*}
where $A \in \R^{n \times d}$ can be treated as a length-$n$ document and each word has length-$d$ embedding size. Here $D = \diag(  A X A^\top {\bf 1}_n )$. For any given $A \in \R^{n \times d}$ and $B \in \R^{n \times n}$, the goal is to find some weight $X$ to optimize the above objective function.
\end{definition}
In contrast to the formulation in \cite{zhdk23,as23,bsz23}, the parameter $X$ in Definition~\ref{def:attention_regression} is equivalent to the $QK^\top \in \R^{d \times d}$ in the generalized version of \cite{zhdk23,as23,bsz23} (e.g. replacing $Q\in \R^{n \times d}$ by $X Q$ where $X \in \R^{n \times d}$ 
and $Q \in \R^{d \times d}$. Similarly for $K$ and $V$. In such scenario, $X$ can be viewed as a matrix representation of a length-$n$ sentence.). A number of work \cite{asa+22,gtlv22,onr+22} study the in-context learning from mathematical perspective in a much simplified setting than Definition~\ref{def:attention_regression}, which is linear regression formulation (see following).
\begin{definition}\label{def:linear_regression}
Given a matrix $A \in \R^{n \times d}$ and $b \in \R^n$, the goal is to solve
\begin{align*}
\min_{x} \| A x - b \|_2
\end{align*}
\end{definition}
Several theoretical transformer work have studied either exponential regression \cite{gms23,lsz23} or softmax regression problem \cite{dls23}. In this work, to take a more step forward to understand the softmax unit in the attention scheme in LLMs. We consider the following softmax regression and study the in-context learning phenomena based on it.
\begin{definition}[Softmax Regression]\label{def:softmax_regression}
Given a $A \in \R^{n \times d}$ and a vector $b \in \R^n$, the goal is to solve
\begin{align*}
\min_{x \in \R^d} \| \langle \exp(Ax) , {\bf 1}_n \rangle^{-1} \exp(Ax) - b \|_2
\end{align*}
\end{definition}
We remark that the Definition~\ref{def:softmax_regression} is a formulation is between Definition~\ref{def:linear_regression} and Definition~\ref{def:attention_regression}.

\subsection{Our Result}
We state our major result as follows:
\begin{theorem}[Bounded shift for Learning in-context, informal of combination of Theorem~\ref{thm:main_formal:x} and Theorem~\ref{thm:main_formal:A}]\label{thm:main_informal}
If the following conditions hold
\begin{itemize}
    \item Let $A \in \R^{n \times d}$.
    \item Let $b \in \R^n$.
    \item $\| A \| \leq R$.
    \item Let $\| x \|_2 \leq R$.
    \item $\| A (x_{t+1} - x_t) \|_{\infty} < 0.01$.
    \item $\| (A_{t+1} - A_t) x \|_{\infty} < 0.01$.
    \item Let $R \geq 4$.
    \item Let $M:=  n^{1.5} \exp(10R^2)$.
\end{itemize}
We consider the softmax regression (Definition~\ref{def:softmax_regression}) problem
\begin{align*}
    \min_x \| \langle \exp(Ax) , {\bf 1}_n \rangle^{-1} \exp(Ax) - b \|_2.
\end{align*}
\begin{itemize}
\item {\bf Part 1.} If we move the $x_t$ to $x_{t+1}$, then we're solving a new softmax regression problem with
\begin{align*}
    \min_x \| \langle \exp(Ax) , {\bf 1}_n \rangle^{-1} \exp(Ax) - \wt{b} \|_2
\end{align*}
where 
\begin{align*}
    \| \wt{b} - b \|_2 \leq M \cdot \| x_{t+1} - x_t \|_2
\end{align*}
\item {\bf Part 2.} If we move the $A_t$ to $A_{t+1}$, then we're solving a new softmax regression with
\begin{align*}
     \min_x \| \langle \exp(Ax) , {\bf 1}_n \rangle^{-1} \exp(Ax) - \wh{b} \|_2
\end{align*}
where 
\begin{align*}
    \| \wh{b} - b \|_2 \leq M \cdot \| A_{t+1} - A_t \|
\end{align*}
\end{itemize}
\end{theorem}

Recall that $A \in \R^{n \times d}$ denotes a length-$n$ document and each word has the length-$d$ embedding size and $x$ denotes the simplified version of $QK^\top$.
One-step gradient descent can be treated as an update to the model's weight $x$.
Thus, part 1 of our result (Theorem \ref{thm:main_informal}) implies that the data transformation of $b$ induced by 
 gradient-descent on the $\ell_2$ regression loss is bounded by $M \cdot \| x_{t+1} - x_t \|_2$.
Following \cite{onr+22}, to do in-context learning, a self-attention layer update can be treated as an update to the tokenized document $A$. Thus, part 2 of our result (Theorem \ref{thm:main_informal}) implies that the data transformation of $b$ induced by  a single self-attention layer is bounded  by $M \cdot \| A_{t+1} - A_t \|$.

We remark that the data transformation of $b$ induced by $1)$ a single self-attention layer and by $2)$ gradient-descent on the $\ell_2$ regression loss are both bounded.
This bounded transformation for the $b$ implies that when training self-attention-only Transformers for fundamental regression tasks, the models
learned by gradient-descent and Transformers show great similarity.

\paragraph{Roadmap.}

In Section \ref{sec:related_work}, we introduce some related work.
In Section \ref{sec:pre}, we give some preliminaries. 
In Section \ref{sec:softmax_x}, we compute the gradient of the loss function with softmax function with respect to $x$. Those functions include $\alpha(x)^{-1}$, $\alpha(x)$ and $f(x)$.
In Section \ref{sec:lipschitz_x}, we show the Lipschitz for the self-attention function with respect to $x$. 
In Section \ref{sec:softmax_a}, we compute the gradient of the loss function with softmax function with respect to $A$.
In Section \ref{sec:lipschitz_a}, we show the Lipschitz for the self-attention function with respect to $A$. 
In Section \ref{sec:main}, we give the main result of our paper.

\section{Related Work}
\label{sec:related_work}

\subsection{In-context Learning}

\cite{asa+22}  indicated that Transformer-based in-context learners are able to perform traditional learning algorithms implicitly. 
This is achieved by encoding smaller models within their internal activations. These smaller models are updated by the given context. 
They theoretically investigate the learning algorithms that Transformer decoders can implement. 
They demonstrate that Transformers need only a limited number of layers and hidden units to implement various linear regression algorithms. 
For $d$-dimensional regression problems, a $O(d)$-hidden-size Transformer can perform a single step of gradient descent.
They also demonstrate that the Transformer with $O(d^2)$ hidden size is able to update a ridge regression problem.
The study reveals that Transformers theoretically have the ability to perform multiple linear regression algorithms. 

 \cite{gtlv22} concentrate on training Transformer to learn certain functions, under in-context conditions. The goal is to have a more comprehensive understanding of in-context learning and determine if Transformers can learn the majority of functions within a given class after training. They found that in-context learning is possible even when there is a distribution shift between the training and inference data or between in-context examples and query inputs.
In addition, they find out that Transformers can learn more complex function classes such as sparse linear functions, two-layer neural networks, and decision trees. These trained Transformers have comparable performance to task-specific learning algorithms.

\cite{onr+22} demonstrate and provide an explanation of the similarity between the training process of the Transformers in in-context tasks and some meta-learning formulations based on gradient descent.
During the process of training Transformers for auto-regressive tasks, the implementation of in-context learning in the Transformer forward pass is carried out through gradient-based optimization of an implicit auto-regressive inner loss that is constructed from the in-context data.

Formally speaking, they consider the following problem $\min_{x} \| A x - b \|_2$ defined in Definition \ref{def:linear_regression}. They first show that doing one step of gradient descent carries out data transformation as follows:
\begin{align*}
    \| A (x + \delta_x) - b \|_2 = & ~ \|A x - (b - \delta_b) \|_2  \\
    = & ~ \|A x - \wt{b} \|_2 
\end{align*}
where $\delta_x$ denotes the one-step gradient descent on $x$ and $\delta_b$ denotes the corresponding data transformation on $b$.
They also show that a self-attention layer is in principle capable of exploiting statistics in the current training data samples. Concretely, let $Q,K,V \in \R^{ d \times d}$ denotes the weights for the query matrix, key matrix, and value matrix respectively.  The linear self-attention layer updates an input sample by doing the following data transformation:
\begin{align*}
    \hat{b}_j = b_j + PVK^\top Q_j 
\end{align*}
where $\hat{b}$ denotes the updated $b$ and  $P$ denotes the projection matrix such that a Transformer step $\hat{b}_j$ on every $j$ is identical to the gradient-induced dynamics $\wt{b}_j$.
This equivalence implies that when training linear-self-attention-only Transformers for fundamental regression tasks, the models learned by GD and Transformers show great similarity.

\cite{xrlm21} explores the occurrence of in-context learning during pre-training when documents exhibit long-range coherence. The Language Model (LLM) develops the ability to generate coherent next tokens by deducing a latent document-level concept. During testing, in-context learning is observed when the LLM deduces a shared latent concept between examples in a prompt. Through the research conducted, it has been demonstrated that in-context learning happens even when there is a distribution mismatch between prompts and pretraining data, especially in scenarios where the pretraining distribution is a mixture of Hidden Markov Models \cite{bp66}.
Theoretically, they show that the error of the in-context predictor is optimal when a distinguishability condition holds. 
In cases where this condition does not hold, the expected error still reduces as the length of each example increases. This finding highlights the importance of both input and input-output mapping contributes to in-context learning.

\subsection{Transformer Theory}

The advancements of Transformers have been noteworthy, however, their learning mechanisms are not completely comprehensible yet. Although these models have performed remarkably well in structured and reasoning activities, our comprehension of their mathematical foundations lags significantly behind.
Past research has indicated that the outstanding performance of Transformer-based models can be attributed to the information within their components, such as the multi-head attention. Various studies \cite{tdp19,  vb19, hl19, b22} have presented empirical proof that these components carry a substantial amount of information, which can help in resolving different probing tasks.

Recent research has investigated the potential of Transformers through both theoretical and experimental methods, including Turing completeness \cite{bpg20}, function approximation \cite{ybr+20, cdw+21}, formal language representation \cite{bag20, egz20, yppn21}, and abstract algebraic operation learning \cite{zbb+22}. Some of these studies have indicated that Transformers may act as universal approximators for sequence-to-sequence operations and emulate Turing machines \cite{pmb19, bpg20}. \cite{lwd+23} demonstrate the existence of contextual sparsity in LLM, which can be accurately predicted. They exploit the sparsity to speed up LLM inference without degrading the performance from both a theoretical perspective and an empirical perspective. 
\cite{dcl+21} proposed the Pixelated Butterfly model that uses a simple fixed sparsity pattern to speed up the training of Transformer. Other studies have focused on the expressiveness of attention within Transformers \cite{dgv+18, vbc20, zkv+20, egkz21, szks21, wcm21}. 

Furthermore, \cite{zpga23} has demonstrated that moderately sized masked language models may effectively parse and recognize syntactic information that helps in the partial reconstruction of a parse tree.   Inspired by the language grammar model studied by \cite{zpga23}, \cite{dgs23} consider the tensor cycle rank approximation problem.  \cite{gms23} consider the exponential regression in neural tangent kernel over-parameterization setting. 
\cite{lsz23} studied the computation of regularized version of the exponential regression problem but they ignore the normalization factor. \cite{dls23} consider the softmax regression which considers the normalization factor compared to exponential regression problems \cite{gms23,lsz23}. The majority of LLMs can perform attention computations in an approximate manner during the inference process, as long as there are sufficient guarantees of precision. This perspective has been studied by various research, including \cite{cgrs19,kkll20,wlk+20,dkod20,kvpf20,cdw+21,cdl+22}. With this in mind, \cite{zhdk23,as23,bsz23,dms23} have conducted a study on the computation of the attention matrix from the hardness perspective and developed faster algorithms.

\section{Preliminary}\label{sec:pre}
In Section \ref{sec:notations}, we introduce the notations used in this paper.
In Section \ref{sec:basicalg}, we give some facts about the basic algebra.
In Section \ref{sec:lowerbound}, we propose the lower bound on $\langle \exp(Ax), {\bf 1}_n \rangle$.

\subsection{Notations}
\label{sec:notations}

For a positive integer $n$, we use $[n]$ to denote $\{1,2,\cdots,n\}$, for any positive integer $n$.

We use $\E[\cdot]$ to denote expectation. We use $\Pr[\cdot]$ to denote probability.

We use ${\bf 1}_n$ to denote the vector where all entries are one. We use ${\bf 0}_0$ to denote the vector where all entries are zero. 
The identity matrix of size $n\times n$ is represented by $I_n$ for a positive integer $n$. 

The symbol $\R$ refers to real numbers and $\R_{\geq 0}$ represents non-negative real numbers.

For any vector $x \in \R^n$, $\exp(x) \in \R^n$ denotes a vector where the $i$-th entry is $\exp(x_i)$ and $\| x \|_2$ represents its $\ell_2$ norm, that is, $\| x \|_2 := ( \sum_{i=1}^n x_i^2 )^{1/2}$. We use $\| x \|_{\infty}$ to denote $\max_{i \in [n]} |x_i|$.

For any vector $x \in \R^n$ and vector $y \in \R^d$, we use $\langle x,y \rangle$ to denote the inner product of vector $x$ and $y$.

The notation $B_i$ is used to indicate the $i$-th row of matrix $B$. 

If $a$ and $b$ are two column vectors in $\R^n$, then $a \circ b$ denotes a column vector where $(a\circ b)_i = a_ib_i$. 

 For a square and full rank matrix $B$, we use $B^{-1}$ to denote the true inverse of $B$.

\subsection{Basic Algebras}
\label{sec:basicalg}

\begin{fact}\label{fac:vector_norm}
For vectors $x,y \in \R^n$, we have
\begin{itemize}
    \item $\| x \circ y \|_2 \leq \| x \|_{\infty} \cdot \| y \|_2$
    \item $\| x \|_{\infty} \leq \| x \|_2 \leq \sqrt{n} \| x \|_{\infty}$
    \item $\| \exp(x) \|_{\infty} \leq \exp(\| x \|_2)$
    \item For any $\| x - y \|_{\infty} \leq 0.01$, we have $\| \exp(x) - \exp(y) \|_2 \leq \| \exp(x) \|_2 \cdot 2 \| x - y \|_{\infty}$
\end{itemize}
\end{fact}

\begin{fact}\label{fac:matrix_norm}
For matrices $X,Y$, we have 
\begin{itemize}
    \item $\| X^\top \| = \| X \|$
    \item $\| X \| \geq \| Y \| - \| X - Y \|$
    \item $\| X + Y \| \leq \| X \| + \| Y \|$
    \item $\| X \cdot Y \| \leq \| X \| \cdot \| Y \|$ 
    \item If $X \preceq \alpha \cdot Y$, then $\| X \| \leq \alpha \cdot \| Y \|$
\end{itemize}
\end{fact}

\subsection{Lower bound on \texorpdfstring{$\beta$}{}}
\label{sec:lowerbound}

\begin{lemma}\label{lem:beta}
If the following conditions holds
\begin{itemize}
    \item $\| A \| \leq R$
    \item $\| x \|_2 \leq R$
    \item Let $\beta$ be lower bound on $\langle \exp(Ax), {\bf 1}_n \rangle$
\end{itemize}
Then we have 
\begin{align*}
    \beta \geq \exp(-R^2)
\end{align*}
\end{lemma}
\begin{proof}
We have
\begin{align*}
\langle \exp(Ax) ,{\bf 1}_n \rangle 
= & ~ \sum_{i=1}^n \exp( (Ax)_i ) \\
\geq & ~ \min_{i \in [n]} \exp( (Ax)_i ) \\
\geq & ~ \min_{i \in [n]} \exp( - | (Ax)_i| ) \\
= & ~ \exp( - \max_{i \in [n]} | (Ax)_i | ) \\
= & ~ \exp(-\| A x \|_{\infty}) \\
\geq & ~ \exp(-\| A x \|_2) \\
\geq & ~ \exp(-R^2)
\end{align*}
the 1st step follows from simple algebra, the 2nd step comes from simple algebra, the 3rd step follows from the fact that $\exp(x) \geq \exp(-|x|)$, the 4th step follows from the fact that $\exp(-x)$ is monotonically decreasing, the 5th step comes from definition of $\ell_{\infty}$ norm, the 6th step follows from Fact~\ref{fac:vector_norm}, the 7th step follows from the assumption on $A$
 and $x$.
\end{proof}

\section{Softmax Function with respect to \texorpdfstring{$x$}{}}\label{sec:softmax_x}
In Section \ref{sec:softmax_x:definition}, we give the definitions used in the computation.
In Section \ref{sec:softmax_x:gradient}, we compute the gradient of the loss function with softmax function with respect to $x$. Those functions includes $\alpha(x)^{-1}$, $\alpha(x)$ and $f(x)$.

\subsection{Definitions}\label{sec:softmax_x:definition}

We define function softmax $f$ as follows
\begin{definition}[Function $f$, Definition 5.1 in \cite{dls23}]\label{def:f:x}
Given a matrix $A \in \R^{n \times d}$. Let ${\bf 1}_n$ denote a length-$n$ vector that all entries are ones.  
We define prediction function $f: \R^d \rightarrow \R^n$ as follows 
\begin{align*}
f(x) := \langle \exp(Ax) , {\bf 1}_n \rangle^{-1} \cdot \exp(Ax) .
\end{align*}
\end{definition}

\begin{definition}[Loss function $L_{\exp}$, Definition 5.3 in \cite{dls23}]\label{def:L_exp:x}
Given a matrix $A \in \R^{n \times d}$ and a vector $b \in \R^n$. 
We define loss function $L_{\exp} : \R^d \rightarrow \R$ as follows 
\begin{align*}
L_{\exp}(x) := 0.5 \cdot \| \langle \exp(Ax) , {\bf 1}_n \rangle^{-1} \exp(A x) - b \|_2^2.
\end{align*}
\end{definition}

For convenient, we define two helpful notations $\alpha$ and $c$
\begin{definition}[Normalized coefficients, Definition 5.4 in \cite{dls23}]
\label{def:alpha:x}
    We define $\alpha : \R^d \rightarrow \R$ as follows
    \begin{align*}
    \alpha(x) := \langle \exp(Ax), {\bf 1}_n \rangle.
    \end{align*}
    Then, we can rewrite $f(x)$ (see Definition~\ref{def:f:x}) and $L_{\exp}(x)$ (see Definition~\ref{def:L_exp:x}) as follows
    \begin{itemize}
        \item $f(x) = \alpha(x)^{-1} \cdot \exp(Ax)$.
        \item $L_{\exp}(x) = 0.5 \cdot \| \alpha(x)^{-1} \cdot \exp(Ax) - b \|_2^2$.
        \item $L_{\exp}(x) = 0.5 \cdot \| f(x) - b \|_2^2$.
    \end{itemize}
\end{definition}

\begin{definition}[Definition 5.5 in \cite{dls23}]
\label{def:c:x}
    We define function $c: \R^d \in \R^n$ as follows
    \begin{align*}
    c(x) := f(x) - b.
    \end{align*}
    Then we can rewrite $L_{\exp}(x)$ (see Definition~\ref{def:L_exp:x}) as follows
    \begin{itemize}
        \item $L_{\exp}(x) = 0.5 \cdot \| c(x) \|_2^2$.
    \end{itemize}
\end{definition}

\subsection{Gradient Computations}\label{sec:softmax_x:gradient}

We state a lemma from previous work,
\begin{lemma}[Gradient, Lemma 5.6 in \cite{dls23}]\label{lem:gradient_computation}
If the following conditions hold
\begin{itemize}
    \item Given matrix $A \in \R^{n \times d}$ and a vector $b \in \R^n$.
    \item Let $\alpha(x)$ be defined in Definition~\ref{def:alpha:x}.
    \item Let $f(x)$ be defined in Definition~\ref{def:f:x}.
    \item Let $c(x)$ be defined in Definition~\ref{def:c:x}.
    \item Let $L_{\exp}(x)$ be defined in Definition~\ref{def:L_exp:x}.
\end{itemize}
 For each $i \in [d]$, we have
\begin{itemize}
    \item Part 1. 
    \begin{align*}
        \frac{ \d \exp(Ax)}{ \d x_i} = \exp(Ax) \circ A_{*,i}
    \end{align*}
    \item Part 2.
    \begin{align*}
        \frac{\d \langle \exp(Ax) , {\bf 1}_n \rangle }{\d x_i} = \langle \exp(Ax) , A_{*,i}\rangle
    \end{align*}
    \item Part 3.
    \begin{align*}
        \frac{\d \alpha(x)^{-1} }{\d x_i} = -\alpha(x)^{-1} \cdot \langle f(x), A_{*,i} \rangle
    \end{align*}
    \item Part 4.
    \begin{align*}
        \frac{\d f(x) }{\d x_i} 
        = \frac{ \d c(x) }{ \d x_i } =  & ~ -  \langle f(x), A_{*,i}\rangle \cdot f(x) + f(x) \circ A_{*,i}
    \end{align*}   
    \item Part 5.  
    \begin{align*}
        \frac{\d L_{\exp} (x) }{\d x_i} =  \underbrace{ A_{*,i}^\top }_{1 \times n} \cdot \Big( \underbrace{f(x) }_{n \times 1} \underbrace{ \langle c(x), f(x)  \rangle }_{ \mathrm{scalar} } + \underbrace{ \diag(f(x)) }_{n \times n} \underbrace{ c(x) }_{n \times 1} \Big)
    \end{align*}
\end{itemize}
\end{lemma}

\section{Lipschitz with respect to \texorpdfstring{$x$}{}}\label{sec:lipschitz_x}
In Section \ref{sec:lipschitz_x:preli}, we give the preliminary to compute the Lipschitz.
In Section \ref{sec:lipschitz_x:upperbound_delta_b}, we show the upper bound of $\delta_b$.
In Section \ref{sec:lipschitz_x:lipschitz_expAx}, we compute the Lipschitiz of function $\exp(Ax)$ with respect to $x$.
In Section \ref{sec:lipschitz_x:lipschitz_alpha}, we compute the Lipschitiz of the function $\alpha$ with respect to $x$.
In Section \ref{sec:lipschitz_x:lipschitz_alpha_inverse}, we compute the Lipschitiz of function $\alpha^{-1}$ with respect to $x$.

\subsection{Preliminary}\label{sec:lipschitz_x:preli}
We can compute 
\begin{align*}
\frac{\d L}{ \d x} = g(x)
\end{align*}

Let $\eta > 0$ denote the learning rate.

We update 
\begin{align*}
x_{t+1} = x_t + \eta \cdot g(x_t)
\end{align*}

\begin{definition}\label{def:delta_b}
We define $\delta_b \in \R^n$ to be the vector that satisfies the following conditions
\begin{align*}
\| \langle \exp( A x_{t+1}), {\bf 1}_n \rangle^{-1} \exp(A x_{t+1} ) - b \|_2^2 = \| \langle \exp(Ax_t) , {\bf 1}_n\rangle^{-1} \exp(Ax_t) - b + \delta_b \|_2^2
\end{align*}
\end{definition}
Let $\{-1,+1\}^n$ denote a vector that each entry can be either $-1$ or $+1$. 
In the worst case, there are $2^n$ possible solutions, e.g.,
\begin{align*}
(\langle \exp( A x_{t+1}), {\bf 1}_n \rangle^{-1} \exp(A x_{t+1} ) - \langle \exp(Ax_t) , {\bf 1}_n\rangle^{-1} \exp(Ax_t)) \circ \{-1,+1\}^n
\end{align*}
The norm of all the choices are the same. Thus, it is sufficient to only consider one solution as follows.
\begin{claim}\label{cla:delta_b}
We can write $\delta_b$ as follows

\begin{align*}
\delta_b = \underbrace{ \langle \exp( A x_{t+1}), {\bf 1}_n \rangle^{-1} \exp(A x_{t+1} ) }_{ f(x_{t+1}) } - \underbrace{ \langle \exp(Ax_t) , {\bf 1}_n\rangle^{-1} \exp(Ax_t) }_{ f(x_t) } .
\end{align*}
\end{claim}
\begin{proof}
The proof directly follows from Definition~\ref{def:delta_b}.
\end{proof}

For convenience, we split $\delta_b$ into two terms, and provide the following definitions
\begin{definition}\label{def:delta_b_1_delta_b_2}
We define
\begin{align*}
\delta_{b,1} := & ~ (\langle \exp(Ax_{t+1}) , {\bf 1}_n \rangle^{-1} - \langle \exp(Ax_{t}) , {\bf 1}_n \rangle^{-1} ) \cdot \exp(A x_{t+1} ) \\
\delta_{b,2} := & ~  \langle \exp(Ax_{t}) , {\bf 1}_n \rangle^{-1} \cdot ( \exp(A x_{t+1} ) - \exp(Ax_t) ) 
\end{align*}
\end{definition}
Thus, we have

\begin{lemma}\label{lem:rewrite_deltab}
We have
\begin{itemize}
    \item \begin{align*}
\delta_b = \delta_{b,1} + \delta_{b,2}
\end{align*}
\item We can rewrite $\delta_{b,1}$ as follows
\begin{align*}
\delta_{b,1} = (  \alpha(x_{t+1})^{-1} - \alpha(x_t)^{-1} ) \cdot \exp( A x_{t+1}) , 
\end{align*}
\item We can rewrite $\delta_{b,2}$ as follows
\begin{align*}
\delta_{b,2} = \alpha(x_{t})^{-1} \cdot ( \exp(A x_{t+1}) - \exp(Ax_t) ). 
\end{align*}
\end{itemize}
\end{lemma}
\begin{proof}
We have
\begin{align*}
    \delta_b = & ~  \delta_{b,1} + \delta_{b,2} \\
    = & ~   \alpha(x_{t+1})^{-1}\exp( A x_{t+1})  - \alpha(x_t)^{-1}\exp( A x_{t+1})  +\\
    & ~   \alpha(x_{t})^{-1} \exp(A x_{t+1}) - \alpha(x_{t})^{-1} \exp(Ax_t)  \\
    = & ~  \alpha(x_{t+1})^{-1}\exp( A x_{t+1}) - \alpha(x_{t})^{-1} \exp(Ax_t) \\
    = & ~ \langle \exp( A x_{t+1}), {\bf 1}_n \rangle^{-1} \exp(A x_{t+1} ) - \langle \exp(Ax_t) , {\bf 1}_n\rangle^{-1} \exp(Ax_t) ,
\end{align*}
where the 1st step follows from the definitions  of $\delta_b$, the 2nd step follows from the definitions of $\delta_{b,1}$ and $\delta_{b,2}$, the 3rd step follows from simple algebra, the 4th step comes from the definition of $\alpha$. 
\end{proof}

\subsection{Upper Bounding \texorpdfstring{$\delta_b$}{} with respect to \texorpdfstring{$x$}{}}\label{sec:lipschitz_x:upperbound_delta_b}
We can show that
\begin{lemma}\label{lem:delta_b_1_delta_b_2:x}
If the following conditions hold
\begin{itemize}
    \item Let $\beta \in (0, 1)$.
    \item Let $\delta_{b,1} \in \R^n$ be defined as Definition~\ref{def:delta_b_1_delta_b_2}.
    \item Let $\delta_{b,2} \in \R^n$ be defined as Definition~\ref{def:delta_b_1_delta_b_2}.
    \item Let $\delta_b = \delta_{b,1} + \delta_{b,2}$.
    \item Let $R \geq 4$.
\end{itemize}
We have
\begin{itemize}
\item Part 1. 
\begin{align*}
    \| \delta_{b,1} \|_2 \leq 2 \beta^{-2} n^{1.5} \exp(2R^2) \cdot \| x_{t+1} - x_t \|_2
\end{align*}
\item Part 2.
\begin{align*}
     \| \delta_{b,2} \|_2 \leq 2 \beta^{-1} \sqrt{n} R \exp(R^2) \cdot \| x_{t+1} - x_t \|_2
\end{align*}
\item Part 3.
\begin{align*}
    \| \underbrace{ f(x_{t+1}) - f(x_t) }_{\delta_b} \|_2 \leq 4 \beta^{-2} n^{1.5} R \exp(2R^2) \cdot \| x_{t+1} - x_t \|_2
\end{align*}
\end{itemize}
\end{lemma}
\begin{proof}
{\bf Proof of Part 1.}
We have
\begin{align*}
\|  \delta_{b,1} \|_2
\leq & ~ |  \alpha(x_{t+1})^{-1} - \alpha(x_t)^{-1} | \cdot \| \exp( A x_{t+1}) \|_2 \\
\leq & ~ |  \alpha(x_{t+1})^{-1} - \alpha(x_t)^{-1} | \cdot \sqrt{n} \cdot \exp(R^2) \\
\leq & ~ \beta^{-2} \cdot | \alpha(x_{t+1} ) - \alpha(x_t) | \cdot \sqrt{n} \cdot \exp(R^2) \\
\leq & ~ \beta^{-2} \cdot \sqrt{n} \cdot \| \exp(Ax_{t+1}) - \exp(A x_t) \|_2 \cdot \sqrt{n} \cdot \exp(R^2) \\
\leq & ~ \beta^{-2} \cdot \sqrt{n} \cdot 2\sqrt{n} R \exp(R^2) \| x_{t+1} - x_t \|_2 \cdot \sqrt{n} \cdot \exp(R^2) \\
= & ~ 2 \beta^{-2} n^{1.5} R \exp(2R^2) \cdot \| x_{t+1} - x_t \|_2
\end{align*}
where the first step follows from definition, the second step follows from assumption on $A$ and $x$, the third step follows Lemma~\ref{lem:lipschitz_alpha_inverse:x}, the forth step follows from Lemma~\ref{lem:lipschitz_alpha:x}, the fifth step follows from Lemma~\ref{lem:lipschitz_exp:x}.

{\bf Proof of Part 2.}

We have
\begin{align*}
\| \delta_{b,2} \|_2 
\leq & ~ | \alpha(x_{t+1})^{-1} | \cdot \|  \exp(A x_{t+1}) - \exp(Ax_t) \|_2 \\
\leq & ~ \beta^{-1} \cdot \|  \exp(A x_{t+1}) - \exp(Ax_t) \|_2 \\
\leq & ~ \beta^{-1} \cdot 2 \sqrt{n} R \exp(2R^2) \cdot \| x_{t+1} - x_t \|_2
\end{align*}
where the first step follows from definition, the 2nd step comes from Lemma \ref{lem:lipschitz_exp:x}.

{\bf Proof of Part 3.}

  We have
  \begin{align*}
      \| \delta_b \|_ 2  = & ~ \| \delta_{b,1} + \delta_{b,2} \|_2  \\
      \leq & ~ \| \delta_{b,1}\|_2 +  \|\delta_{b,2} \|_2 \\
      \leq & ~ 2 \beta^{-2} n^{1.5} R \exp(2R^2) \cdot \| x_{t+1} - x_t \|_2 + 2 \beta^{-1}   n^{0.5} R \exp(2R^2) \cdot \| x_{t+1} - x_t \|_2 \\
      \leq & ~ 2 \beta^{-2} n^{1.5} R \exp(2R^2) \cdot \| x_{t+1} - x_t \|_2 + 2  \beta^{-2} n^{1.5} R \exp(2R^2) \cdot \| x_{t+1} - x_t \|_2 \\
       \leq & ~ 4 \beta^{-2} n^{1.5} R \exp(2R^2) \cdot \| x_{t+1} - x_t \|_2
  \end{align*}
  where the 1st step follows from the definition of $\delta_b$,
  the 2nd step follows from triangle inequality, 
  the 3rd step follows from the results in Part 1 and Part 2,
  the 4th step follows from the fact that $ n  \geq 1 $ and $\beta^{-1} \geq 1$, 
  the 5th step follows from simple algebra.
\end{proof}

\subsection{Lipschitz for function \texorpdfstring{$\exp(Ax)$}{} with respect to \texorpdfstring{$x$}{}}\label{sec:lipschitz_x:lipschitz_expAx}

\begin{lemma}\label{lem:lipschitz_exp:x}
If the following conditions holds
\begin{itemize}
    \item Let $A \in \R^{n \times d}$
    \item Let $\| A (y-x) \|_{\infty} < 0.01$
    \item Let $\| A \| \leq R$
    \item Let $x, y$ satisfy that $\| x \|_2 \leq R$ and $\| y \|_2 \leq R$
\end{itemize}
Then we have
\begin{align*}
    \| \exp(Ax) - \exp(Ay )\|_2 \leq 2 \sqrt{n} R \exp(R^2) \cdot \| x - y \|_2.
\end{align*}
\end{lemma}
\begin{proof}
We have
\begin{align*}
\| \exp(Ax) - \exp(Ay) \|_2 
\leq & ~ \| \exp(Ax) \|_2 \cdot 2 \| A (x-y) \|_{\infty} \notag \\
\leq & ~ \sqrt{n} \cdot \exp(\| A x\|_2)  \cdot 2 \| A (x-y) \|_{\infty} \notag \\
\leq & ~ \sqrt{n} \exp(R^2) \cdot 2 \| A (x-y) \|_2 \notag \\
\leq & ~ \sqrt{n} \exp(R^2)  \cdot 2 \| A \| \cdot \| x - y \|_2 \notag\\
\leq & ~ 2 \sqrt{n} R \exp(R^2) \cdot \|x - y\|_2
\end{align*}
where the 1st step follows  from $\| A (y-x) \|_{\infty} < 0.01$ and Fact~\ref{fac:vector_norm}, 
the 2nd step comes from Fact~\ref{fac:vector_norm}, 
the 3rd step follows from Fact~\ref{fac:matrix_norm}, 
the 4th step follows from Fact~\ref{fac:matrix_norm}, 
the last step follows from $\|A\| \leq R$.
\end{proof}

\subsection{Lipschitz for function \texorpdfstring{$\alpha(x)$}{} with respect to \texorpdfstring{$x$}{}}\label{sec:lipschitz_x:lipschitz_alpha}

We state a tool from previous work \cite{dls23}.
\begin{lemma}[Lemma 7.2 in \cite{dls23}]\label{lem:lipschitz_alpha:x}
If the following conditions hold
\begin{itemize}
    \item Let $\alpha(x)$ be defined as Definition~\ref{def:alpha:x}
\end{itemize}
Then we have
\begin{align*}
    | \alpha(x) - \alpha(y) | \leq \| \exp(Ax) - \exp(Ay) \|_2 \cdot \sqrt{n}.
\end{align*}
\end{lemma}

\subsection{Lipschitz for function \texorpdfstring{$\alpha(x)^{-1}$}{} with respect to \texorpdfstring{$x$}{}}\label{sec:lipschitz_x:lipschitz_alpha_inverse}
We state a tool from previous work \cite{dls23}.
\begin{lemma}[Lemma 7.2 in \cite{dls23}]\label{lem:lipschitz_alpha_inverse:x}
If the following conditions hold
\begin{itemize}
    \item Let $\langle \exp(Ax), {\bf 1}_n \rangle \geq \beta$
    \item Let $\langle \exp(Ay), {\bf 1}_n \rangle \geq \beta$
\end{itemize}
Then, we have
\begin{align*}
     | \alpha(x)^{-1} - \alpha(y)^{-1} | \leq \beta^{-2} \cdot | \alpha(x) - \alpha(y) |.
\end{align*}
\end{lemma}

\section{Softmax Function with respect to \texorpdfstring{$A$}{}}\label{sec:softmax_a}

In this section, we consider the function with respect to $A$.
We define function softmax $f$ as follows
\begin{definition}[Function $f$, Reparameterized $x$ by $A$ in Definition~\ref{def:f:x}]\label{def:f:A}
Given a matrix $A \in \R^{n \times d}$. Let ${\bf 1}_n$ denote a length-$n$ vector that all entries are ones.  
We define prediction function $f: \R^{n \times d} \rightarrow \R^n$ as follows 
\begin{align*}
f(A) := \langle \exp(Ax) , {\bf 1}_n \rangle^{-1} \cdot \exp(Ax) .
\end{align*}
\end{definition}

Similarly, we reparameterized $x$ by $A$ for our loss function $L$. We define loss function $L$ as follows
\begin{definition}[Loss function $L_{\exp}$, Reparameterized $x$ by $A$ in Definition~\ref{def:L_exp:x}]\label{def:L_exp:A}
Given a matrix $A \in \R^{n \times d}$ and a vector $b \in \R^{n \times d}$. 
We define loss function $L_{\exp} : \R^{n \times d} \rightarrow \R$ as follows 
\begin{align*}
L_{\exp}(A) := 0.5 \cdot \| \langle \exp(Ax) , {\bf 1}_n \rangle^{-1} \exp(A x) - b \|_2^2.
\end{align*}
\end{definition}

For convenience, we define two helpful notations $\alpha$ and $c$ with respect to $A$ as follows:
\begin{definition}[Normalized coefficients, Reparameterized $x$ by $A$ in Definition~\ref{def:alpha:x}]
\label{def:alpha:A}
    We define $\alpha : \R^{n \times d} \rightarrow \R$ as follows
    \begin{align*}
    \alpha(A) := \langle \exp(Ax), {\bf 1}_n \rangle.
    \end{align*}
    Then, we can rewrite $f(A)$ (see Definition~\ref{def:f:A}) and $L_{\exp}(A)$ (see Definition~\ref{def:L_exp:A}) as follows
    \begin{itemize}
        \item $f(A) = \alpha(A)^{-1} \cdot \exp(Ax)$.
        \item $L_{\exp}(A) = 0.5 \cdot \| \alpha(A)^{-1} \cdot \exp(Ax) - b \|_2^2$.
        \item $L_{\exp}(A) = 0.5 \cdot \| f(A) - b \|_2^2$.
    \end{itemize}
\end{definition}

\begin{definition}[Reparameterized $x$ by $A$ in Definition~\ref{def:c:x}]
\label{def:c:A}
    We define function $c: \R^{n \times d} \in \R^n$ as follows
    \begin{align*}
    c(A) := f(A) - b.
    \end{align*}
    Then we can rewrite $L_{\exp}(A)$ (see Definition~\ref{def:L_exp:A}) as follows
    \begin{itemize}
        \item $L_{\exp}(A) = 0.5 \cdot \| c(A) \|_2^2$.
    \end{itemize}
\end{definition}

\section{Lipschitz with respect to \texorpdfstring{$A$}{}}\label{sec:lipschitz_a}
In Section \ref{sec:lipschitz_a:prelim}, we give the preliminary to compute the Lipschitz.
In Section \ref{sec:lipschitz_a:upperbound_deltab}, we show the upper bound of $\delta_b$ with respect to $A$.
In Section \ref{sec:lipschitz_a:lipschitz_expAx}, we compute the Lipschitiz of function $\exp(Ax)$ with respect to $A$.
In Section \ref{sec:lipschitz_a:lipschitz_alpha}, we compute the Lipschitiz of the function $\alpha$ with respect to $A$.
In Section \ref{sec:lipschitz_a:lipschitz_alpha_inverse}, we compute the Lipschitiz of function $\alpha^{-1}$ with respect to $A$.

\subsection{Preliminary}\label{sec:lipschitz_a:prelim}

We define $\delta_b$ as follows

\begin{definition}[Reparameterized $x$ by $A$ in Definition~\ref{def:delta_b}]\label{def:delta_b:A}
We define $\delta_b \in \R^n$ to be the vector that satisfies the following conditions
\begin{align*}
\| \langle \exp( A_{t+1} x), {\bf 1}_n \rangle^{-1} \exp(A_{t+1} x ) - b \|_2^2 = \| \langle \exp(A_t x) , {\bf 1}_n\rangle^{-1} \exp(A_t x) - b + \delta_b \|_2^2
\end{align*}
\end{definition}

\begin{claim}[Reparameterized $x$ by $A$ in Definition~\ref{cla:delta_b}]
We can write $\delta_b$ as follows
\begin{align*}
\delta_b = \underbrace{ \langle \exp( A_{t+1} x), {\bf 1}_n \rangle^{-1} \exp(A_{t+1} x ) }_{ f(A_{t+1}) } - \underbrace{ \langle \exp(A_t x) , {\bf 1}_n\rangle^{-1} \exp(A_t x) }_{ f(A_t) }.
\end{align*}
\end{claim}
\begin{proof}
The proof directly follows from Definition~\ref{def:delta_b:A}.
\end{proof}

For convenient, we split $\delta_b$ into two terms, and provide the following definitions
\begin{definition}[Reparameterized $x$ by $A$ in Definition~\ref{def:delta_b_1_delta_b_2}]\label{def:delta_b_1_delta_b_2:A}
We define
\begin{align*}
\delta_{b,1} := & ~ (\langle \exp(A_{t+1} x ) , {\bf 1}_n \rangle^{-1} - \langle \exp(A_t x) , {\bf 1}_n \rangle^{-1} ) \cdot \exp(A_{t+1} x ) \\
\delta_{b,2} := & ~  \langle \exp(A_t x) , {\bf 1}_n \rangle^{-1} \cdot ( \exp(A_{t+1} x ) - \exp(A_t x) ) 
\end{align*}
\end{definition}
Thus, we have

\begin{lemma}[Reparameterized $x$ by $A$ in Lemma \ref{lem:rewrite_deltab}]
We have
\begin{itemize} 
    \item We can rewrite $\delta_b \in \R^n$ as follows
    \begin{align*}
\delta_b = \delta_{b,1} + \delta_{b,2}
\end{align*}
\item We can rewrite $\delta_{b,1} \in \R^n$ as follows
\begin{align*}
\delta_{b,1} = (  \alpha(A_{t+1})^{-1} - \alpha(A_t)^{-1} ) \cdot \exp( A_{t+1} x) , 
\end{align*}
\item We can rewrite $\delta_{b,2} \in \R^n$ as follows
\begin{align*}
\delta_{b,2} = \alpha(A_{t})^{-1} \cdot ( \exp(A_{t+1} x) - \exp(A_t x) ). 
\end{align*}
\end{itemize}
\end{lemma}
\begin{proof}
We have
\begin{align*}
    \delta_b = & ~  \delta_{b,1} + \delta_{b,2} \\
    = & ~   \alpha(A_{t+1})^{-1}\exp( A_{t+1} x )  - \alpha(A_t)^{-1}\exp( A_{t+1} x )  +\\
    & ~   \alpha(A_{t})^{-1} \exp(A_{t+1} x ) - \alpha(A_{t})^{-1} \exp(A_t x)  \\
    = & ~  \alpha(A_{t+1})^{-1}\exp( A_{t+1} x ) - \alpha(A_{t})^{-1} \exp(A_t x) \\
    = & ~ \langle \exp( A_{t+1} x), {\bf 1}_n \rangle^{-1} \exp( A_{t+1} x ) - \langle \exp(A_t x) , {\bf 1}_n\rangle^{-1} \exp(A_t x) ,
\end{align*}
where the 1st step follows from the definitions  of $\delta_b$, the 2nd step follows from the definitions of $\delta_{b,1}$ and $\delta_{b,2}$, the 3rd step comes from simple algebra, the 4th step comes from the definition of $\alpha$. 
\end{proof}

\subsection{Upper Bounding \texorpdfstring{$\delta_b$}{} with respect to \texorpdfstring{$A$}{}}\label{sec:lipschitz_a:upperbound_deltab}
We can show that
\begin{lemma}[Reparameterized $x$ by $A$ in Lemma \ref{lem:delta_b_1_delta_b_2:x}]\label{lem:delta_b_1_delta_b_2:A}
If the following conditions hold
\begin{itemize}
    \item Let $\beta \in (0, 1)$.
    \item Let $\delta_{b,1} \in \R^n$ be defined as Definition~\ref{def:delta_b_1_delta_b_2:A}.
    \item Let $\delta_{b,2} \in \R^n$ be defined as Definition~\ref{def:delta_b_1_delta_b_2:A}.
    \item Let $\delta_b = \delta_{b,1} + \delta_{b,2}$.
    \item Let $R \geq 4$.
\end{itemize}
We have
\begin{itemize}
\item Part 1. 
\begin{align*}
    \| \delta_{b,1} \|_2 \leq 2 \beta^{-2} n^{1.5} \exp(2R^2) \cdot \| A_{t+1} - A_t \|_2
\end{align*}
\item Part 2.
\begin{align*}
     \| \delta_{b,2} \|_2 \leq 2 \beta^{-1} \sqrt{n} R \exp(R^2) \cdot \| A_{t+1} - A_t \|_2
\end{align*}
\item Part 3.
\begin{align*}
    \| \underbrace{ f(A_{t+1}) -  f(A_t) }_{ \delta_b } \|_2 \leq 4 \beta^{-2} n^{1.5} R \exp(2R^2) \cdot \| A_{t+1} - A_t \|_2
\end{align*}
\end{itemize}
\end{lemma}
\begin{proof}
{\bf Proof of Part 1.}
We have
\begin{align*}
\|  \delta_{b,1} \|_2
\leq & ~ |  \alpha(A_{t+1})^{-1} - \alpha(A_t)^{-1} | \cdot \| \exp( A_{t+1} x ) \|_2 \\
\leq & ~ |  \alpha(A_{t+1})^{-1} - \alpha(A_t)^{-1} | \cdot \sqrt{n} \cdot \exp(R^2) \\
\leq & ~ \beta^{-2} \cdot | \alpha(A_{t+1} ) - \alpha(A_t) | \cdot \sqrt{n} \cdot \exp(R^2) \\
\leq & ~ \beta^{-2} \cdot \sqrt{n} \cdot \| \exp(A_{t+1} x) - \exp(A_t x) \|_2 \cdot \sqrt{n} \cdot \exp(R^2) \\
\leq & ~ \beta^{-2} \cdot \sqrt{n} \cdot 2\sqrt{n} R \exp(R^2) \| A_{t+1} - A_t \| \cdot \sqrt{n} \cdot \exp(R^2) \\
= & ~ 2 \beta^{-2} n^{1.5} R \exp(2R^2) \cdot \| A_{t+1} - A_t \|
\end{align*}
where the first step follows from definition, the second step follows from assumption on $A$ and $x$, the third step follows Lemma~\ref{lem:lipschitz_alpha_inverse:A}, the forth step follows from Lemma~\ref{lem:lipschitz_alpha:A}, the fifth step follows from Lemma~\ref{lem:lipschitz_exp:A}.

{\bf Proof of Part 2.}

We have
\begin{align*}
\| \delta_{b,2} \|_2 
\leq & ~ | \alpha(A_{t+1})^{-1} | \cdot \|  \exp(A_{t+1} x ) - \exp(A_t x) \|_2 \\
\leq & ~ \beta^{-1} \cdot \|  \exp(A_{t+1} x ) - \exp(A_t x) \|_2 \\
\leq & ~ \beta^{-1} \cdot 2 \sqrt{n} R \exp(2R^2) \cdot \| A_{t+1} - A_t \|
\end{align*}

{\bf Proof of Part 3.}

  We have
  \begin{align*}
      \| \delta_b \|_ 2  = & ~ \| \delta_{b,1} + \delta_{b,2} \|_2  \\
      \leq & ~ \| \delta_{b,1}\|_2 +  \|\delta_{b,2} \|_2 \\
      \leq & ~ 2 \beta^{-2} n^{1.5} R \exp(2R^2) \cdot \| A_{t+1} - A_t \| + 2 \beta^{-1}   n^{0.5} R \exp(2R^2) \cdot \| A_{t+1} - A_t \| \\
      \leq & ~ 2 \beta^{-2} n^{1.5} R \exp(2R^2) \cdot \| A_{t+1} - A_t \| + 2  \beta^{-2} n^{1.5} R \exp(2R^2) \cdot \| A_{t+1} - A_t \| \\
       \leq & ~ 4 \beta^{-2} n^{1.5} R \exp(2R^2) \cdot \| A_{t+1} - A_t \|
  \end{align*}
  where the 1st step follows from the definition of $\delta_b$,
  the 2nd step comes from triangle inequality, 
  the 3rd step comes from the results in Part 1 and Part 2,
  the 4th step follows from the fact that $ n  \geq 1 $ and $\beta^{-1} \geq 1$, 
  the 5th step follows from simple algebra.
\end{proof}

\subsection{Lipschitz for function \texorpdfstring{$\exp(Ax)$}{} with respect to \texorpdfstring{$A$}{}}\label{sec:lipschitz_a:lipschitz_expAx}

\begin{lemma}[Reparameterized $x$ by $A$ in Lemma \ref{lem:lipschitz_exp:x}]\label{lem:lipschitz_exp:A}
If the following conditions holds
\begin{itemize}
    \item Let $A , B \in \R^{n \times d}$
    \item Let $\| (A-B) x \|_{\infty} < 0.01$
    \item Let $\| A \| \leq R$
    \item Let $x$ satisfy that $\| x \|_2 \leq R$
\end{itemize}
Then we have
\begin{align*}
    \| \exp(Ax) - \exp(Bx )\|_2 \leq 2 \sqrt{n} R \exp(R^2) \cdot \| A - B \|.
\end{align*}
\end{lemma}
\begin{proof}
We have
\begin{align*}
\| \exp(Ax) - \exp(Bx) \|_2 
\leq & ~ \| \exp(Ax) \|_2 \cdot 2 \| (A-B) x\|_{\infty} \notag \\
\leq & ~ \sqrt{n} \cdot \exp(\| A x\|_2)  \cdot 2 \| (A-B) x\|_{\infty} \notag \\
\leq & ~ \sqrt{n} \exp(R^2) \cdot 2 \| (A-B) x \|_2 \notag \\
\leq & ~ \sqrt{n} \exp(R^2)  \cdot 2 \| A -B\| \cdot \| x  \|_2 \notag\\
\leq & ~ 2 \sqrt{n} R \exp(R^2) \cdot \|A - B\|
\end{align*}
where the 1st step follows  from $\| A (y-x) \|_{\infty} < 0.01$ and Fact~\ref{fac:vector_norm}, 
the 2nd step follows from Fact~\ref{fac:vector_norm}, 
the 3rd step follows from Fact~\ref{fac:matrix_norm}, 
the 4th step comes from Fact~\ref{fac:matrix_norm}, 
the last step follows from $\|A\| \leq R$.
\end{proof}

\subsection{Lipschitz for function \texorpdfstring{$\alpha(A)$}{} with respect to \texorpdfstring{$A$}{}}\label{sec:lipschitz_a:lipschitz_alpha}

\begin{lemma}[Reparameterized $x$ by $A$ in Lemma \ref{lem:lipschitz_alpha:x}]\label{lem:lipschitz_alpha:A}
If the following conditions hold
\begin{itemize}
    \item Let $\alpha(A)$ be defined as Definition~\ref{def:alpha:A}
\end{itemize}
Then we have
\begin{align*}
    | \alpha(A) - \alpha(B) | \leq \| \exp(Ax) - \exp(Bx) \|_2 \cdot \sqrt{n}.
\end{align*}
\end{lemma}

\begin{proof}
We have
\begin{align*}
| \alpha(A) - \alpha(B)| = & ~ | \langle \exp(Ax) - \exp(Bx) , {\bf 1}_n \rangle | \\
\leq & ~ \| \exp(Ax) - \exp(Bx) \|_2 \cdot \sqrt{n}
\end{align*}
where the 1st step comes from the definition of $\alpha(x)$,
the 2nd step follows from Cauchy-Schwarz inequality (Fact~\ref{fac:vector_norm}).
\end{proof}

\subsection{Lipschitz for function \texorpdfstring{$\alpha(A)^{-1}$}{} with respect to \texorpdfstring{$A$}{}}\label{sec:lipschitz_a:lipschitz_alpha_inverse}

\begin{lemma}[Reparameterized $x$ by $A$ in Lemma \ref{lem:lipschitz_alpha_inverse:x}]\label{lem:lipschitz_alpha_inverse:A}
If the following conditions hold
\begin{itemize}
    \item Let $\langle \exp(Ax), {\bf 1}_n \rangle \geq \beta$
    \item Let $\langle \exp(Bx), {\bf 1}_n \rangle \geq \beta$
\end{itemize}
Then, we have
\begin{align*}
     | \alpha(A)^{-1} - \alpha(B)^{-1} | \leq \beta^{-2} \cdot | \alpha(A) - \alpha(B) |.
\end{align*}
\end{lemma}

\begin{proof}

We can show that
\begin{align*}
| \alpha(A)^{-1} - \alpha(B)^{-1} | 
= & ~ \alpha(A)^{-1} \alpha(B)^{-1} \cdot | \alpha(A) - \alpha(B) | \\
\leq & ~ \beta^{-2} \cdot | \alpha(A) - \alpha(B) |
\end{align*}
where the 1st step follows from simple algebra, 
the 2nd step follows from $\alpha(A) \geq \beta, \alpha(B) \geq \beta$.
\end{proof}

\section{Main Results}\label{sec:main}
In Section \ref{sec:result:wrp_x}, we show our upper bound result of $\delta_b$ with respect to $x$. 
In Section \ref{sec:result:wrp_A}, we show our upper bound result of $\delta_b$ with respect to $A$. 

\subsection{Shifting Weight Parameter \texorpdfstring{$x$}{}}\label{sec:result:wrp_x}
\begin{theorem}[Bounded shift for Learning in-context, informal of Theorem~\ref{thm:main_informal}]\label{thm:main_formal:x}
If the following conditions hold
\begin{itemize}
    \item Let $A \in \R^{n \times d}$
    \item $\| A \| \leq R$
    \item $\| A (x_{t+1} - x_t) \|_{\infty} < 0.01$
    \item Let $R \geq 4$
    \item Let $M:=  n^{1.5} \exp(10R^2)$.
\end{itemize}
We consider the softmax regression problem
\begin{align*}
    \min_x \| \langle \exp(Ax) , {\bf 1}_n \rangle^{-1} \exp(Ax) - b \|_2
\end{align*}
If we move the $x_t$ to $x_{t+1}$, then we're solving a new softmax regression problem with
\begin{align*}
    \min_x \| \langle \exp(Ax) , {\bf 1}_n \rangle^{-1} \exp(Ax) - \wt{b} \|_2
\end{align*}
where 
\begin{align*}
    \| \wt{b} - b \|_2 \leq M \cdot \| x_{t+1} - x_t \|_2
\end{align*}
\end{theorem}
\begin{proof}

We have
\begin{align*}
        \| \wt{b} - b \|_2 \leq  & ~ 4 \beta^{-2} n^{1.5} R \exp(2R^2) \cdot \| x_{t+1} - x_t \|_2 \\
        \leq & ~ 4 n^{1.5} R \exp(2R^2)  \exp(2R^2) \cdot \| x_{t+1} - x_t \|_2 \\
        \leq & ~  n^{1.5} (4R) \exp(4R^2) \cdot \| x_{t+1} - x_t \|_2 \\
        \leq & ~   n^{1.5} \exp(6R^2)\exp(4R^2) \cdot \| x_{t+1} - x_t \|_2 \\
        \leq & ~   n^{1.5} \exp(10 R^2) \cdot \| x_{t+1} - x_t \|_2 \\
        \leq & ~  M \cdot \| x_{t+1} - x_t \|_2 
\end{align*}
where the 1st step follows from Lemma \ref{lem:delta_b_1_delta_b_2:x}, the 2nd step comes from Lemma \ref{lem:beta}, the 3rd step comes from simple algebra, the 4th step follows from simple algebra, the 5th step follows from simple algebra and the 6th step follows from the definition of $M$.
\end{proof}

\subsection{Shifting Sentence Data \texorpdfstring{$A$}{}}\label{sec:result:wrp_A}
\begin{theorem}[Bounded shift for Learning in-context, informal of Theorem~\ref{thm:main_informal}]\label{thm:main_formal:A}
If the following conditions hold
\begin{itemize}
    \item Let $A \in \R^{n \times d}$
    \item $\| A \| \leq R$
    \item $\| ( A_{t+1} -A_t) x \|_{\infty} < 0.01$
    \item Let $R \geq 4$
    \item Let $M:=  n^{1.5} \exp(10R^2)$.
\end{itemize}
We consider the softmax regression problem
\begin{align*}
    \min_x \| \langle \exp(Ax) , {\bf 1}_n \rangle^{-1} \exp(Ax) - b \|_2
\end{align*}
If we move the $x_t$ to $x_{t+1}$, then we're solving a new softmax regression problem with
\begin{align*}
    \min_x \| \langle \exp(Ax) , {\bf 1}_n \rangle^{-1} \exp(Ax) - \wt{b} \|_2
\end{align*}
where 
\begin{align*}
    \| \wt{b} - b \|_2 \leq M \cdot \| A_{t+1} - A_t \|.
\end{align*}
\end{theorem}
\begin{proof}

We have
\begin{align*}
        \| \wt{b} - b \|_2 \leq  & ~ 4 \beta^{-2} n^{1.5} R \exp(2R^2) \cdot \| A_{t+1} - A_t \| \\
        \leq & ~ 4 n^{1.5} R \exp(2R^2)  \exp(2R^2) \cdot \| A_{t+1} - A_t \| \\
        \leq & ~  n^{1.5} (4R) \exp(4R^2) \cdot \| A_{t+1} - A_t \| \\
        \leq & ~   n^{1.5} \exp(6R^2)\exp(4R^2) \cdot \| A_{t+1} - A_t \| \\
        \leq & ~   n^{1.5} \exp(10 R^2) \cdot \| A_{t+1} - A_t \| \\
        \leq & ~  M \cdot \| A_{t+1} - A_t \|
\end{align*}
where the 1st step follows from Lemma \ref{lem:delta_b_1_delta_b_2:x}, the 2nd step follows from Lemma \ref{lem:beta}, the 3rd step follows from simple algebra, the 4th step comes from simple algebra, the 5th step comes from simple algebra and the 6th step follows from the definition of $M$.
\end{proof}

\ifdefined\isarxiv
\bibliographystyle{alpha}
\bibliography{ref}
\else
\bibliography{ref}
\bibliographystyle{alpha}

\fi

\newpage
\onecolumn
\appendix




\end{document}